\documentclass{article}
\usepackage{spconf, amsmath, graphicx}
\usepackage{subcaption, siunitx, booktabs}
\usepackage{bm, amsfonts}
\usepackage{cite}
\usepackage{booktabs}

\def\x{{\bm x}}

\def\z{{\bm z}}

\def\D{{\mathcal D}}

\def\E{{\mathbb E}}

\def \tA{{\text A}}
\def \tN{{\text N}}
\def \tS{{\text S}}
\def \tT{{\text T}}
\def \TheEnc{\bm{\theta}_{{\mathrm {Enc}}}}
\def \TheDec{\bm{\theta}_{{\mathrm {Dec}}}}
\def \TheDom{\bm{\theta}_{{\mathrm {Dom}}}}
\def \LVAE{l_\mathrm{VAE}}
\def \LDOM{l_\mathrm{dom}}
\def \wDOM{w_\mathrm{dom}}
\def \wVAE{w_\mathrm{vae}}
\def \Enc{\mathrm{Enc}}
\def \Dec{\mathrm{Dec}}

\title{Zero-shot domain adaptation of anomalous samples for semi-supervised anomaly detection}
\name{Tomoya Nishida, Takashi Endo, Yohei Kawaguchi} %\thanks{Thanks to XYZ agency for funding.}}
\address{Research and Development Group, Hitachi, Ltd.\\
1-280, Higashi-koigakubo, Kokubunji-shi, Tokyo 185-8601, Japan}
\begin{document}
\ninept
\maketitle
\begin{abstract}
\vspace{-3pt}
Semi-supervised anomaly detection~(SSAD) is a task where normal data and a limited number of anomalous data are available for training.
In practical situations, SSAD methods suffer adapting to domain shifts, since anomalous data are unlikely to be available for the target domain in the training phase.
To solve this problem, we propose a domain adaptation method for SSAD where no anomalous data are available for the target domain.
First, we introduce a domain-adversarial network to a variational auto-encoder-based SSAD model to obtain domain-invariant latent variables.
Since the decoder cannot reconstruct the original data solely from domain-invariant latent variables, we conditioned the decoder on the domain label.
To compensate for the missing anomalous data of the target domain, we introduce an importance sampling-based weighted loss function that approximates the ideal loss function.
Experimental results indicate that the proposed method helps adapt SSAD models to the target domain when no anomalous data are available for the target domain.

%In practical situations, adapting SSAD models to a new data domain~(target domain) is crucial for its usage.
%, since data distribution can change between the training data and the test data.
%However, this adaptation is often infeasible, since anomalous data are likely to be unavailable for the target domain.
%We tackle a domain adaptation problem for semi-supervised anomaly detection.
%semi-supervised anomaly detection is a task where normal data and a limited number of anomalous data are available for training.
%In this task, anomalous data that is the same type as the training data, and anomalous data that is different from the training data are both required to be detected.
%It tries to take advantage of the available anomalous data to detect samples similar to them accurately, while also detecting xxx.
%One issue of semi-supervised anomaly detection is that the distribution of the data to perform anomaly detection, the target data distribution, can be different from that of the initially collected training data.
%Adapting semi-supervised anomaly detection models to a new domain~(target domain) is crucial for its practical usage, since data distribution can change between the training data and the test data.
%However, this adaptation is often infeasible, since anomalous data are likely to be unavailable for the target domain.
%In such situations, anomalous data are often unavailable for the target domain.
%Thus, existing domain adaptation techniques will fail.
% DANN + domain label説明
% importance weighting説明
\end{abstract}
\begin{keywords}
Anomaly detection, domain adaptation, importance weighting
\vspace{-7pt}
\end{keywords}
\section{INTRODUCTION}
\label{sec:intro}
\vspace{-7pt}
Anomaly detection is an essential technology for various business fields such as factory automation~\cite{koizumi2019unsupervised, suefusa2020anomalous} and automated surveillance~\cite{xiao2015learning}.
In these fields, this technology is useful for automatically detecting factory equipment failures or occurrences of unusual events~(e.g., screams or fights).
Typically, anomaly detection is solved as an unsupervised task, where only normal data are available for training.
This is because anomalous data occur in rare situations and are highly diverse, making them difficult to collect.
In this unsupervised setting, well-established methods exist such as methods based on support vector machines~\cite{scholkopf2001estimating}, auto-encoders~\cite{zhou2017anomaly}, or variational auto-encoders~(VAEs)~\cite{an2015variational}.

In some cases, a limited amount of anomalous data may be available for the training data~\cite{gornitz2013toward}.
For example, the target equipment may accidentally break, or an unusual event may occur during data collection. 
In this situation, we can use the available anomalous data to enhance anomaly detection performance.
A notable aspect of this situation is that the collected anomalous data does not necessarily cover all the possible anomalous data, owing to its diversity.
For example, factory equipment can fail in many different ways, where each causes a different kind of anomalous data. 
Here, we are required to detect two types of anomalous data: Anomalous data covered by the collected ones~(seen anomalous data) and those not covered by them~(unseen anomalous data).
Therefore, although anomaly detection aims to classify each data as normal or anomalous, it cannot be solved as a standard binary classification task.
In this paper, we will call this task the semi-supervised anomaly detection~(SSAD) task and focus on it.
SSAD aims to detect seen anomalous data better than unsupervised anomaly detection~(UAD) methods while maintaining the detection performance against unseen anomalous data.  
SSAD methods based on VAEs~\cite{kawachi2018complementary, kawachi2019twoclass}, regression models~\cite{liu2019margin}, or classifiers~\cite{zhang2020video} have been proposed in the literature. 
A similar task was also considered in \cite{ruff2019deep, zhou2022feature}, where they assumed that abundant unlabeled data was also available in addition to the labeled normal and anomalous data.
%This enables us to collect certain types of anomalous data.
%An important aspect of this situation is that other types of anomalous data also exist, owing to the diversity of anomalous data.

% 「異なる状況で異常検知をしなければならない場合がある」ことがちゃんと伝わるように
%The simplest way of accomplishing this in unsupervised anomaly detection is to fine-tune~\cite{} or retrain the models.
%A more faster way of domain adaptation has also been proposed~\cite{}, where it conducts batch normalization to match the data distribution between source data and target data.
In practical situations, the distribution of the data to perform anomaly detection, the target domain data distribution, can be different from that of the initially collected training data, the source domain data distribution.
For instance, the difference in measurement conditions or states of the target equipment may cause this difference.
In such situations, we must adapt the trained model to the target domain.
If both normal and anomalous data are available for the target domain, this can be carried out easily by typical domain adaptation methods~\cite{ganin2016domain, Yang2020anomaly} or by retraining the model.
However, anomalous data for the target domain are often unavailable since they rarely occur.
In this case, adaptations of SSAD models will fail since only normal data will be available for the target domain.
%In this case, only normal data will be available for the target domain.
%This will cause the adaptation of the semi-supervised anomaly detection model to fail.
%Thus, even though some anomalous data are available in the source domain, SSAD methods cannot be carried out in the target domain.

%While some investigations for domain adaptation of unsupervised anomaly detection methods exist, domain adaptation for semi-supervised anomaly detection has not been investigated.
%If both normal data and appropriate anomalous data can be collected for the target data, adaptation can still be conducted by fine-tuning.
%\item However, collecting the appropriate anomaly data for the target data is often difficult. 
%Here, the appropriate anomalous data is the same type of anomalous data as the initially collected anomaly data.
%However, collecting this appropriate anomaly data may be impossible, since the same cause of the anomaly is unlikely to happen again.
%In this case, only normal data will be available for the target domain.
%This will cause the adaptation of the semi-supervised anomaly detection model to fail.
% 代わりに教師なし異常検知なら実行できるが，ソースドメインで入手した異常データを全く活用できなくなってしまう。

In this paper, we propose a domain adaptation method for SSAD, where normal data and seen anomalous data are available for the source domain, but only normal data is available for the target domain. 
Under this situation, we aim to detect seen anomalous data in the target domain more accurately than UAD methods while maintaining the detection performance against unseen anomalous data.
First, we add a domain-adversarial network to the VAE-based SSAD model to obtain domain-invariant latent variables.
Since the decoder of the VAE cannot reconstruct the original data solely from domain-invariant latent variables, we condition the decoder with the domain label.
To compensate for the missing target anomalous data, we approximate the ideal loss function with an importance sampling-based weighted loss function~\cite{ishii2020partially} for the available data.
By this method, the trained model can detect seen anomalous data better than UAD models even when those data are unavailable for the target domain.

\section{RELATED WORK}
\label{sec:related}
\vspace{-7pt}
Several works on domain adaptation for anomaly detection exist in the literature.
For example, in \cite{yamaguchi2019adaflow}, Normalizing Flow was unified with Adaptive Batch-Normalization to conduct domain adaptation of an UAD model.
Domain adaptation for SSAD was also considered in \cite{zhang2020video}, where the SSAD model was composed of a supervised anomaly detection model and an UAD model.
However, in this work, only the UAD model was adapted to the target domain.
Therefore, the adaptation does not help to detect seen anomalous data in the target domain.

If we look at tasks other than anomaly detection, domain adaptation has been widely investigated for classification tasks.
Typically, domain adaptation methods aim to match the distributions between the embedded features of the source data and that of the target data~\cite{ganin2016domain, long2015learning}.
Complicated tasks have also been tackled, such as partial domain adaptation~\cite{zhang2018importance, cao2018partial}.
Here, a specific subset of classes will be absent in the target domain data, which is a similar problem setting to ours.
However, in this task, the missing classes will not appear in the test data and therefore do not need to be correctly discriminated.
In comparison, in our problem setting, data from the missing class, which is the seen anomalous data, will appear in the test data and needs to be correctly detected.
The closest task is tackled in \cite{ishii2020partially}, which assumes that several classes are missing in the target training data, while all classes appear in the test data.
Compared to this task, we assume that data missing from both the source and target domain, the unseen anomalous data, will also appear in the test data. 
%Thus, this problem setting has not been considered in the literature.

\section{PROBLEM STATEMENT}
\label{sec:prob}
\vspace{-7pt}
\begin{table}[!t]
    \centering
    \caption{Data availability in our problem setting.}
    \vspace{-6pt}
    \begin{tabular}{@{}c|c c c@{}}
         & Normal & Seen anomaly & Unseen anomaly \\
        \hline
        Source & Yes & Yes & No\\
        Target & Yes & No & No
    \end{tabular}
    \label{tab:data}
    \vspace{-15pt}
\end{table}

In this paper, we tackle the problem of domain adaptation for SSAD.
Table.~\ref{tab:data} shows what types of data are available for the training phase.
For the source domain, normal data and some anomalous data are available. 
We will call the available anomalous data the "seen" anomalous data.
For the target domain, only normal data is available.
Thus, the overall training dataset is summarized as
\begin{align}
    \D = \left\{(\x, c, d) \mid \x \in \mathbb{R}^s, (c, d) \in \{(\tN, \tS), (\tA, \tS), (\tN, \tT)\} \right\},
\end{align}
where $\x \in R^s$ is the input data, $c$ denotes whether the input data belongs to the normal data~($c=\tN$) or the seen anomalous data~($c=\tA$), and $d$ denotes the domain of the input data; $d=\tS$ for source domain data and $d=\tT$ for target domain data.
We also assume that anomalous data that are different from the seen anomalous data also exist in the test data. 
We will call this the "unseen" anomalous data.

In this paper, we aim to create an anomaly detection model that can detect seen anomalous data more accurately than UAD methods while maintaining the detection performance against unseen anomalous data in the target domain.
Note that no seen anomalous data is available for the target domain.
Hence, we cannot train an SSAD model straightforwardly.
Moreover, adapting a model that was trained in the source domain to the target domain by standard domain adaptation techniques will fail.

\section{PROPOSED METHOD}
\label{sec:prop}
\vspace{-7pt}
\subsection{Two-class VAE for SSAD}
\label{ssec:da}
\vspace{-5pt}
As the baseline method, we introduce the SSAD framework proposed in \cite{kawachi2018complementary, kawachi2019twoclass}.
Here, a VAE is trained with two prior distributions $p\left( \z \mid c=\tN \right)$ and $p\left( \z \mid c=\tA \right)$ of the latent variable $\z$.
Each prior distribution corresponds to the distribution of the normal and anomalous data.
By setting the two prior distributions far from each other, we can easily discriminate seen anomalous data from normal data by observing the latent variable.
We will call this model the two-class VAE~(2C-VAE).
The loss function of a 2C-VAE is defined as
\begin{align}
    %&
    \LVAE
    \left(
        \x, c; \TheEnc, \TheDec
    \right)
    %\notag \\
    &=
    \E_{q_{\TheEnc}\left( \z \mid \x \right)}
    \left[
        \log p_{\TheDec} \left(\x \mid \z \right)
    \right]
    \notag \\
    &~~~~~~-
    {\text K}{\text L}
    \left[
        q_{\TheEnc} \left(\z \mid \x \right)
        \|
        p \left(\z \mid c \right)
    \right],
    \label{eq: 2c-vae loss}
\end{align}
which is the evidence lower bound~(ELBO) of $\log p\left( \x \right)$.
%As noted above, the prior distribution of the latent variable $p \left(\z \mid c \right)$ depends on whether the data is normal or anomalous.
Here, $q_{\TheEnc}\left(\z|\x\right)$ and $p_{\TheDec}\left(\x \mid \z\right)$ are constructed by neural networks.
$q_{\TheEnc}\left(\z|\x\right)$ is constructed by an encoder $f_{Enc}$.
This encoder outputs the parameters $\bm{u}$ of a probabilistic density function $p_\Enc(\cdot;\bm{u})$, which will be used as $q_{\TheEnc}\left(\z|\x\right)$, as
\begin{align}
    & \bm{u} = f_\Enc\left( \x ; \TheEnc \right),
    \\
    & q_{\TheEnc}\left(\z \mid \x\right) = p_\Enc\left(\z ; \bm{u}\right).
\end{align}
For $p_\Enc$, Gaussian distributions~\cite{kawachi2018complementary} or von Mises-Fishier~(vMF) distributions~\cite{kawachi2019twoclass} are used.
$p_{\TheDec}\left(\x \mid \z\right)$ is constructed by a decoder $f_\Dec$.
This decoder typically outputs the expected value of $x$, and a Gaussian distribution around it is used to form $p_{\TheDec}\left(\x \mid \z\right)$, as
\begin{align}
    & \tilde{\x} = f_\Dec\left( \z; \TheDec \right),
    \\
    & p_{\TheDec}\left(\x \mid \z\right) = \mathcal{N}\left(\x ; \tilde{\x}, \bm{I}\right).
\end{align}
Thus, $\log p_{\TheDec} \left(\x \mid \z \right)$ can be calculated as the mean squared error between $\tilde{\x}$ and $\x$.

%$p\left( \z \mid c \right)$ is the prior distribution of the latent variable which depends on whether the data is normal or anomalous. 
%The two prior distributions $p\left( \z \mid c=N \right)$ and $p\left( \z \mid c=A \right)$ are set to be far from each other. 
%By this, the seen anomalous data can be easily discriminated from the normal data by observing the latent variable.
% complementary, sphericalみたいな話入れる。

\subsection{Domain-adversarial network for 2C-VAE}
\vspace{-5pt}
\begin{figure}
    %TODO:本文にないパラメタ値消す 
    \centering
    \includegraphics[width=6cm]{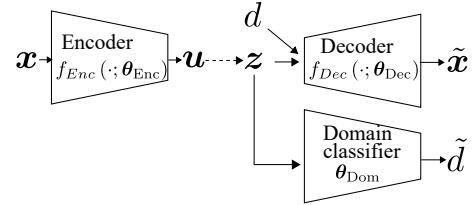}
    \vspace{-7pt}
    \caption{Model architecture of the proposed method}
    \label{fig: model}
    \vspace{-15pt}
\end{figure}
To adapt the 2C-VAE to the target domain, we introduce a domain classifier $p_{\TheDom} \left( d \mid \z \right)$ to construct an adversarial network~\cite{ganin2016domain}.
The domain classification loss is written as
\begin{align}
    &
    \LDOM
    \left(
        \x, d; \TheEnc, \TheDom
    \right)
    =
    \E_{q_{\TheEnc}\left( \z \mid \x \right)}
    \left[
        \log p_{\TheDom} \left( d \mid \z \right)
    \right].
    \label{eq: domain loss}
\end{align}
Through adversarial training of \eqref{eq: domain loss}, we aim to match the distributions between $\z$ of the source data and that of the target data, i.e., we aim to obtain a domain-invariant latent variable.
However, this aim contradicts the optimization of \eqref{eq: 2c-vae loss} since the decoder cannot reconstruct the original data without any domain information.
To overcome this problem, we input the domain label to the decoder in addition to the latent variable.
By using the domain label, the decoder can estimate the original data even when $\z$ becomes domain-invariant.
% TODO: sourceとtargetドメインでLVの特徴量が一致するのを促進するために用いる。ところが，ドメイン情報が無くなった特徴量からデコーダで元のデータを復元するのは困難である。そこで，2C-VAEのデコーダには，LVの他に真のドメインラベルもone-hot vectorで与える。

By combining \eqref{eq: domain loss} with \eqref{eq: 2c-vae loss}, the optimization problem for training the 2C-VAE is written as
%loss function for a given sample $(\x, c, d)$ is written as
\begin{align}
    &
    \min_{\TheEnc, \TheDec}
    \max_{\TheDom}
    \mathbb{E}
    \left[
    \mathcal{L}
    \left(
        \x, c, d;\TheEnc, \TheDec, \TheDom
    \right)
    \right],
    \label{eq: opt prob}
    \\
    &
    \mathcal{L}
    \left(
        \x, c, d;\TheEnc, \TheDec, \TheDom
    \right)
    \notag \\
    &:=
    -
    \LVAE
    \left(
        \x, c, d; \TheEnc, \TheDec
    \right)
    +
    \lambda_{dom}
    \LDOM
    \left(
        \x, d; \TheEnc, \TheDom
    \right), \label{eq: loss fnc}
\end{align}
where $l_{dom} > 0$ is a hyperparameter.
Note that we added the domain label $d$ to $\LVAE$ since it is now input to the decoder.
Here, $\TheEnc$ and $\TheDec$ are trained to minimize $\mathcal{L}$, whereas $\TheDom$ is trained to maximize $\mathcal{L}$.
Thus, $\TheDom$ is trained to correctly classify the domain from $\z$, while $\TheEnc$ is trained to fool the domain classifier.
By this, the model will obtain a latent variable that is domain-invariant.
We summarize the model architecture in Fig.~\ref{fig: model}. 
From here, we will omit writing down the parameters in the loss function when it is obvious.

When training the model with \eqref{eq: loss fnc}, the absence of seen anomalous data in the target domain causes a problem.
In \ref{ssec:vae loss} and \ref{ssec:domain loss}, we will describe this problem and introduce an importance sampling-based method~\cite{ishii2020partially} to overcome it.

\vspace{-10pt}
\subsection{Approximation of 2C-VAE loss}
\label{ssec:vae loss}
\vspace{-5pt}
The expected value of $\LVAE$ can be decomposed as
\begin{align}
    &
    \E \left[ 
        \LVAE
        \left(
            \x, c, d; \TheEnc, \TheDec
        \right)
    \right]
    \notag \\
    &=
    \int 
    \LVAE
    \left(
        \x, c=\tN, d=\tS
    \right)
    p \left( \x, c=\tN, d=\tS \right)
    d\x
    \notag \\
    &+
    \int 
    \LVAE
    \left(
        \x, c=\tA, d=\tS
    \right)
    p \left( \x, c=\tA, d=\tS \right)
    d\x
    \notag \\
    &+
    \int 
    \LVAE
    \left(
        \x, c=\tN, d=\tT
    \right)
    p \left( \x, c=\tN, d=\tT \right)
    d\x
    \notag \\
    &+
    \int 
    \LVAE
    \left(
        \x, c=\tA, d=\tT
    \right)
    p \left( \x, c=\tA, d=\tT \right)
    d\x.
    \label{eq: vae loss original}
\end{align}
Since we have no anomalous samples from the target domain, data sampled from $p \left( \x, c=\tA, d=\tT \right)$ are unavailable.
Thus, the fourth term cannot be calculated from the available data.
This means that if we optimize $l_{2CVAE}\left( \x, c, d \right)$ using the available data, the latent variable of the seen anomalous data from the target domain will not be restricted to the prior distribution $p\left(\z | c=\tA \right)$.
Hence, detecting seen anomalous data from the target domain will fail.

To overcome this problem, we rewrite $p \left( \x, c=\tA, d=\tT \right)$ as
\begin{align}
    p \left( \x, c=\tA, d=\tT \right)
    &=
    p \left( \x, c=\tA, d=\tS \right)
    \frac{p \left( \x, c=\tA, d=\tT \right)}{p \left( \x, c=\tA, d=\tS \right)} %\label{eq: reweighting dom}
    %\notag \\
    %&=
    %p \left( \x, c=\tA, d=\tS \right)
    %\cdot
    %\frac{
    %    p \left( d=\tT \mid \x, c=\tA \right)
    %}{
    %    p \left( d=\tS \mid \x, c=\tA \right)
    %}
    \notag \\
    &=
    p \left( \x, c=\tA, d=\tS \right)
    %\cdot
    \frac{
        p \left( d=\tT \mid \x \right)
    }{
        p \left( d=\tS \mid \x \right)
    }. \label{eq: w_vae}
\end{align}
%The second term of \eqref{eq: reweighting dom} can be reformulated as
%\begin{align}
%    \frac{
%        p \left( \x, c=\tA, d=\tT \right)
%    }{
%        p \left( \x, c=\tA, d=\tS \right)
%    }
%    &=
%    \frac{
%        p \left( d=\tT \mid \x, c=\tA \right) p\left( x, c=\tA \right)
%    }{
%        p \left( d=\tS \mid \x, c=\tA \right) p \left( x, c=\tA \right)
%    }
%    \notag \\
%    &=
%    \frac{
%        p \left( d=\tT \mid \x, c=\tA \right)
%    }{
%        p \left( d=\tS \mid \x, c=\tA \right)
%    }
%    \notag \\
%    &=
%    \frac{
%        p \left( d=\tT \mid \x \right)
%    }{
%        p \left( d=\tS \mid \x \right)
%    }
%    := w_{VAE}.
%    \label{eq: w_vae}
%\end{align}
Here, we assumed
$p \left( d \mid \x, c=\tA \right) = p \left( d \mid \x, c=\tN \right) = p \left( d \mid \x \right)$.
This can be interpreted as the covariate shift condition under class shifts instead of the standard covariate shift condition under domain shifts. 
This is a reasonable assumption when the domain of the data can be estimated only from the data without knowing whether it is normal or anomalous.

Now, we can use the source anomalous data with a weight expressed in the second term of \eqref{eq: w_vae} to calculate the fourth term of \eqref{eq: vae loss original}.
Since we are training the domain classifier along with the 2C-VAE, the second term of \eqref{eq: w_vae} can be estimated as
\begin{align}
    \wVAE
    &:=
    \frac{
        p \left( d=\tT \mid \x \right)
    }{
        p \left( d=\tS \mid \x \right)
    }
    =
    \frac{
        \int q_{\TheEnc} \left( \z \mid \x \right) p_{\TheDom} \left( d=\tT \mid \z \right) d\z
    }{
        \int q_{\TheEnc} \left( \z \mid \x \right) p_{\TheDom} \left( d=\tS \mid \z \right) d\z
    }.
\end{align}

Here, we assumed $p\left(d \mid \z, \x \right) = p\left(d \mid \z \right)$, which is an appropriate assumption since the trained domain classifier can often estimate the domain label accurately only from $\z$ in practical cases.
In addition to the reformulation described above, we set $p(d=\tS) = p(d=\tT) = 0.5$ and
$p\left( c \mid d=\tT \right) = p\left( c \mid d=\tS \right) := p_c$,
where $p\left( c \mid d=\tS \right)$ is calculated from the source data.
Finally, \eqref{eq: vae loss original} can be reformulated as
\begin{align}
    &
    \E \left[ 
        \LVAE
        \left(
            \x, c; \TheEnc, \TheDec
        \right)
    \right]
    \notag \\
    &=
    \frac{p_{\tN}}{2}
    \int 
    \LVAE
    \left(
        \x, c=\tN, d=\tS
    \right)
    % \notag \\
    % & \qquad \qquad \qquad \qquad \qquad \qquad \qquad
    p \left( \x \mid c=\tN, d=\tS \right)
    d\x
    \notag \\
    &+
    \frac{p_{\tA}}{2}
    \int 
    \left(
    \LVAE
    \left(
        \x, c=\tA, d=\tS
    \right)
    \right.
    \notag \\
    &\left.
    \qquad ~~%\qquad \qquad \qquad
    +
    \wVAE
    ~
    \LVAE
    \left(
        \x, c=\tA, d=\tT
    \right)
    \right)
    %\notag \\
    %& \qquad \qquad \qquad \qquad \qquad \qquad \qquad
    p \left( \x \mid c=\tA, d=\tS \right)
    d\x
    \notag \\
    &+
    \frac{p_{\tN}}{2}
    \int 
    \LVAE
    \left(
        \x, c=\tN, d=\tT
    \right)
    %\notag \\
    %& \qquad \qquad \qquad \qquad \qquad \qquad \qquad
    p \left( \x \mid c=\tN, d=\tT \right)
    d\x.
    \label{eq: weighted classification loss}
\end{align}
Here, all terms can be calculated by the available data.
Note that in the second term, the data $x$ is sampled from the source anomalous data, but in $\LVAE\left(\x, c=\tA, d=\tT \right)$ the domain label is $d=\tT$. 
Thus, we input $d=\tT$ to the decoder to calculate this term, although the data $\x$ will be sampled from the source domain.
As a result, \eqref{eq: weighted classification loss} applies large weights to data from missing classes if it is regarded to be similar to target domain data.
This is different from partial domain adaptation problems, where importance sampling was used to apply small weights to data from missing classes [18].

\vspace{-5pt}
\subsection{Approximation of domain classification loss}
\label{ssec:domain loss}
\vspace{-5pt}
\begin{table}[t]
    \centering
    \caption{Experimental conditions}
    \vspace{-10pt}
    \resizebox{0.4\textwidth}{!}{
    \begin{tabular}{@{} cccc @{}}
        \toprule
         & Normal & Seen anomaly & Unseen anomaly \\
        \midrule
        Case 1 & 1, 2, 3 & 4, 5, 6 & 7, 8, 9 \\ 
        Case 2 & 4, 5, 6 & 7, 8, 9 & 1, 2, 3 \\
        Case 3 & 7, 8, 9 & 1, 2, 3 & 4, 5, 6 \\
        \bottomrule
    \end{tabular}
    }
    \vspace{-15pt}
    \label{tab: exp cond}
\end{table}

\begin{table*}[t]
    \centering
    \caption{AUC values for seen anomalous data~(\%)}
    \vspace{-10pt}
    \resizebox{0.8\textwidth}{!}{
    \begin{tabular}{@{} r ccc ccc ccc ccc c @{}} 
        \toprule
        Method & \multicolumn{3}{c}{Prop. w/ weights} & \multicolumn{3}{c}{Prop. w/o weights} &  \multicolumn{3}{c}{2C-vMF-VAE-da} & \multicolumn{3}{c}{2C-vMF-VAE} & VAE\\
        \cmidrule(lr){2-4}\cmidrule(lr){5-7}\cmidrule(lr){8-10}\cmidrule(lr){11-13}\cmidrule(l){14-14}
        Value used for anomaly score & RL & KL & ELBO & RL & KL & ELBO & RL & KL & ELBO & RL & KL & ELBO & RL\\ 
        \midrule
        Case 1 & 88.2 & 92.9 & $\bm{97.4}$ & 92.5 & 90.1 & 96.8 & 80.7 & 92.2 & 94.6 & 79.5 & 90.8 & 95.3 & 95.9 \\
        Case 2 & 79.3 & 75.6 & $\bm{88.6}$ & 84.9 & 71.7 & 86.9 & 73.3 & 74.9 & 87.0 & 77.8 & 76.5 & 87.3 & 85.5 \\
        Case 3 & 85.2 & 77.7 & $\bm{92.7}$ & 86.8 & 70.5 & 91.8 & 79.7 & 67.7 & 88.4 & 79.8 & 69.0 & 90.5 & 82.3 \\
        Avg.   & 84.2 & 82.1 & $\bm{92.9}$ & 88.1 & 77.4 & 91.8 & 77.9 & 78.3 & 90.0 & 79.0 & 78.8 & 91.0 & 87.9 \\
        \hline
    \end{tabular}
    }
    \label{tab: result seen}
\end{table*}

\begin{table*}[t]
    \centering
    \caption{AUC values for unseen anomalous data~(\%)}
    \vspace{-10pt}
    \resizebox{0.8\textwidth}{!}{
    \begin{tabular}{@{} r ccc ccc ccc ccc c @{}}
        \toprule
        Method & \multicolumn{3}{c}{Prop. w/ weights} & \multicolumn{3}{c}{Prop. w/o weights} &  \multicolumn{3}{c}{2C-vMF-VAE-da} & \multicolumn{3}{c}{2C-vMF-VAE} & VAE\\
        \cmidrule(lr){2-4}\cmidrule(lr){5-7}\cmidrule(lr){8-10}\cmidrule(lr){11-13}\cmidrule(l){14-14}
        Value used for anomaly score & RL & KL & ELBO & RL & KL & ELBO & RL & KL & ELBO & RL & KL & ELBO & RL\\ 
        \midrule
        Case 1 & 83.7 & 72.7 & 89.7 & 88.5 & 70.4 & 90.2 & 77.4 & 79.7 & 86.0 & 77.5 & 77.8 & 86.5 & $\bm{90.6}$ \\
        Case 2 & 80.0 & 78.2 & 89.1 & 87.8 & 76.3 & $\bm{90.1}$ & 76.6 & 84.1 & 87.6 & 79.8 & 85.9 & 89.0 & 88.8 \\
        Case 3 & 86.4 & 60.4 & 86.8 & 86.8 & 60.9 & 87.1 & 81.9 & 58.5 & 83.9 & 84.8 & 60.6 & 85.5 & $\bm{90.0}$ \\
        Avg. & 83.4 & 70.4 & 88.5 & 87.7 & 69.2 & 89.1 & 78.6 & 74.1 & 85.8 & 80.7 & 74.8 & 87.0 & $\bm{89.8}$ \\
        \hline
    \end{tabular}
    }
    \label{tab: result unseen}
    \vspace{-15pt}
\end{table*}

%The same problem seen in \eqref{eq: vae loss original} occurs for the domain classification loss.
\eqref{eq: weighted classification loss} will be an accurate estimate of \eqref{eq: vae loss original} if $\wVAE$ is accurate.
However, the domain classifier that calculates the components of $\wVAE$ also cannot be trained accurately because of the unavailable data.
To overcome this problem, we will also reformulate the domain classification loss.
First, we rewrite $p \left( \x, c=\tA, d=\tT \right)$ in the same procedure as in $\eqref{eq: w_vae}$, as
\begin{align}
    p \left( \x, c=\tA, d=\tT \right)
    &=
    p \left( \x, c=\tN, d=\tT \right)
    \frac{p \left( \x, c=\tA, d=\tT \right)}{p \left( \x, c=\tN, d=\tT \right)}
    \notag \\
    &=
    p \left( \x, c=\tN, d=\tT \right)
    \frac{
        p \left( c=\tA \mid \x \right)
    }{
        p \left( c=\tN \mid \x \right)
    }
    \notag \\
    &:=
    p \left( \x, c=\tN, d=\tT \right)
    \wDOM.
    \label{eq: reweighting class}
\end{align}
%Here, by following the same procedure as in \eqref{eq: w_vae}, the second term of \eqref{eq: reweighting class} can be rewritten as
%\begin{align}
%    &
%    \frac{
%        p \left( \x, c=\tA, d=\tT \right)
%    }{
%        p \left( \x, c=\tN, d=\tT \right)
%    }
%    =
%    \frac{
%        p \left( c=\tA \mid \x \right)
%    }{
%        p \left( c=\tN \mid \x \right)
%    }
%    := w_{dom}. \label{eq: w_domain}
%\end{align}
Here, we assumed the covariate shift condition
$p\left(c | \x, d=\tT \right) = p\left(c | \x, d=\tS \right) = p\left( c | \x \right)$.

Unlike in \cite{ishii2020partially}, $\wDOM$ cannot be calculated directly from the model outputs since we do not explicitly classify the given sample as normal or anomalous.
However, by assuming $p(\x \mid c, \z) = p(\x \mid \z)$, the numerator and the denominator of $\wDOM$ can be calculated as
\begin{align}
    &
    p \left( c=\tA \mid \x \right)
    =
    \int 
        q_{\TheEnc}\left( \z \mid \x \right) 
        \frac{
            p_{\tA}p\left( \z \mid c=\tA \right)
        }{
            p\left(\z\right)
        }
    d\z,
    \\
    &
    p \left( c=\tN \mid \x \right)
    =
    \int 
    q_{\TheEnc}\left( \z \mid \x \right) 
    \frac{
        p_{\tN} p\left( \z \mid c=\tN \right)
    }{
        p\left(\z\right)
    }
    d\z,
    \\
    &
    p\left(\z\right)
    =
    p_{\tN}~p\left( \z \mid c=\tN \right) + p_{\tA}~p\left( \z \mid c=\tA \right).
\end{align}

Finally, the domain classification loss can be reformulated as
\begin{align}
    &
    \E \left[ 
        \LDOM
        \left(
            \x, d; \TheEnc, \TheDom
        \right)
    \right]
    \notag \\
    &=
    \frac{p_{\tN}}{2}
    \int 
    \LDOM
    \left(
        \x, d=\tS
    \right)
    % \notag \\
    % & \qquad \qquad \qquad \qquad \qquad \qquad \qquad
    p \left( \x \mid c=\tN, d=\tS \right)
    d\x
    \notag \\
    &+
    \frac{p_{\tA}}{2}
    \int 
    \LDOM
    \left(
        \x, d=\tS
    \right)
    % \notag \\
    % & \qquad \qquad \qquad \qquad \qquad \qquad \qquad
    p \left( \x \mid c=\tA, d=\tS \right)
    d\x
    \notag \\
    &+
    \frac{p_{\tN}}{2}
    \int 
    \left( 1 + \wDOM \right)
    \LDOM
    \left(
        \x, d=\tT
    \right)
    % \notag \\
    % & \qquad \qquad \qquad \qquad \qquad \qquad \qquad
    p \left( \x \mid c=\tN, d=\tT \right)
    d\x.
    \label{eq: weighted domain loss}
\end{align}

Now, the overall optimization problem for domain adaptation is formulated by substituting \eqref{eq: weighted classification loss} and \eqref{eq: weighted domain loss} to \eqref{eq: opt prob}.

\section{EXPERIMENTS}
\label{sec:exp}
\vspace{-7pt}

\subsection{Experimental conditions}
\label{ssec:exp conditions}
\vspace{-5pt}

% データの説明
To investigate the effectiveness of the proposed method, we conducted an SSAD experiment under domain shifts.
We used two datasets of handwritten digit images: MNIST~\cite{deng2012mnist} for the source domain and rotated MNIST~\cite{ghifary2015domain} for the target domain. 
The rotated MNIST is the MNIST dataset rotated for 45 degrees.
Similarly to \cite{kawachi2019twoclass}, we divided the digits into three classes to create an SSAD task: Normal, seen anomaly, and unseen anomaly.
We examined three cases summarized in Table~\ref{tab: exp cond}.
For training, we assumed that the normal and seen anomalous data are available for the source domain and that only normal data are available for the target domain.
%The goal is to detect both seen and unseen anomalous data in the target domain.
Here, the proposed method aims to detect seen anomalous data in the target domain more accurately than UAD methods while maintaining the detection performance against unseen anomalous data.

% 各手法の説明
For the original model of the proposed method~(Prop. w/ weights), we used the two-class vMF-VAE~\cite{kawachi2019twoclass}.
As explained in Sect.~4, we added a domain classifier to this original model and used the domain label as an additional input of the decoder.
To evaluate the effectiveness of importance weighting, we also conducted the proposed method with the importance weights set to $0$, i.e., $w_{vae}=w_{dom}=0$~(Prop. w/o weights).
For the conventional method, we evaluated the VAE-based UAD method~(VAE) and the original 2C-vMF-VAE-based SSAD method~(2C-vMF-VAE).
We trained VAE with the normal data from the target domain and 2C-vMF-VAE with all the available data, i.e., the normal and seen anomalous data of the source domain and the normal data of the target domain.
We also evaluated a method where a standard domain adaptation technique, DANN~\cite{ganin2016domain} was integrated into the original SSAD method~(2C-vMF-VAE-da).
Here, we simply introduced a domain classifier to the 2C-vMF-VAE model and conducted domain-adversarial learning.
%Therefore, the difference between Prop. w/o weights and 2C-vMF-VAE-da is that, in 2C-vMF-VAE-da, the domain label is not input to the decoder.
For VAE, we used the reconstruction loss~(RL) as the anomaly score.
For other methods, we used the RL, the KL divergence loss from $p(\z \mid c=\tN)$~(KL), and the ELBO as the anomaly score.
For the encoder and decoder of each model, we used a single hidden linear layer with $50$ units.
For the domain classifier, we also used a single hidden linear layer with $50$ units.

% 学習方法の説明
We trained all models by Adam~\cite{Kingma2014}. 
VAE was trained for $200$ epochs with the learning rate set to $0.01$. 
All the other models were trained for $500$ epochs with the learning rate set to $0.0001$.
For models that require a domain classifier, we set $\lambda_{dom} = 0.01$.
For the proposed method, we regularized the values of $\wDOM$ and $\wVAE$ to be
$\tilde{w} = 1/2 \min\{w, 2\}$, where $w \in \{\wDOM, \wVAE\}$.
This was necessary since the values could be unstable at the early stage of training.

\vspace{-5pt}
\subsection{Results}
\label{ssec: results}
\vspace{-5pt}
We compared the area under the receiver operating characteristic curve~(AUC) scores for anomaly detection against the target domain data.
Here, the scores were calculated separately for seen anomalous data and unseen anomalous data.

Table~\ref{tab: result seen} shows the AUC values of detecting seen anomalous data in the target domain. 
For all methods except VAE, using ELBO as the anomaly score achieved higher AUC values than RL or KL.
A possible reason is because some seen anomalous samples in the target domain were relatively similar to that in the source domain, while others were not.
In this case, samples similar to the source domain samples can be detected by KL, while other samples can be detected by RL.
Therefore, both types of anomalous samples can be detected by combining the two scores into ELBO, which achieves the best performance.
Because of this, we will mainly focus on the AUC values achieved by ELBO.
First, 2C-vMF-VAE-da showed lower AUC values than 2C-vMF-VAE.
This result implies that simply applying domain adaptation will completely fail when seen anomalous data is unavailable for the target domain.
In contrast, the proposed method~(Prop. w/ weights) outperformed all other methods, including the VAE-based UAD method.
This result suggests that the proposed method successfully adapted the model to the target domain in this situation.
In addition, the fact that the proposed method showed higher AUC values than the method without weights~(Prop. w/o weights) shows the effectiveness of the importance weights.
%Prop. w/ weights showed the highest AUC values when we used ELBO as the anomaly score. 
%This shows the effectiveness of introducing the importance weights and giving the domain label to the decoder.
%Note that 2C-vMF-VAE-da showed lower AUC values than 2C-vMF-VAE. 
%This result implies that simply applying adversarial domain adaptation will fail when seen anomalous data is unavailable for the target domain.
%In comparison, the proposed method successfully adapted the model to the target domain in this situation.
%For all methods except VAE, using ELBO as the anomaly score showed higher AUC values than using RL or KL. 
%This can be because some seen anomalous samples in target domain were relatively easy to adapt from the source data, while others were difficult.
%In this case, the easy-to-adapt samples can be detected by KL, while difficult-to-adapt samples can be detected by RL.
%Thus, both types of anomalous data can be detected by combining these scores to ELBO, which achieves the best performance.

Table~\ref{tab: result unseen} shows the AUC values of detecting unseen anomalous data in the target domain.
Here, VAE showed the highest AUC values in two cases. 
Still, the difference between the AUC values of VAE and those of the proposed method was relatively small.
Also, the proposed method even beat VAE in one case.
Therefore, the proposed method mostly maintained the detection performance against unseen anomalous data.

Overall, the proposed method improved the detection performance against seen anomalous data when it was unavailable in the target domain while maintaining the detection performance against unseen anomalous data compared to the unsupervised method. 
Thus, the proposed method enabled domain adaptation of the SSAD model when anomalous data was missing for the target domain.

\vspace{-7pt}
\section{CONCLUSION}
\label{sec:conclusion}
\vspace{-7pt}
We proposed a domain adaptation method for SSAD where no anomalous data are available for the target domain.
First, we conducted domain-adversarial training on the VAE-based SSAD model to obtain domain-invariant latent variables.
Since the decoder cannot reconstruct the original data solely from a domain-invariant latent variable, we conditioned the decoder with the domain label.
To compensate for the unavailable data, we introduced an importance sampling-based weighted loss function that approximates the ideal loss function. 
Experimental results showed that the proposed method detects seen anomalous data better than UAD while maintaining the detection performance against unseen anomalous data.
This result indicates that the proposed method helps adapt SSAD models to the target domain when no anomalous data are available for the target domain.

%\vfill\pagebreak
\vfill\pagebreak
\bibliographystyle{IEEEbib}
\bibliography{str_def_abrv, refs}

\end{document}